\documentclass[runningheads]{llncs}
\usepackage[T1]{fontenc}
\usepackage{graphicx}
\usepackage{amsmath}
\usepackage{amssymb}
\usepackage{booktabs}
\usepackage{multirow}
\usepackage{makecell} 

\newcommand{\norm}[1]{\left\lVert#1\right\rVert}

\begin{document}
\title{Explainable Deep Convolutional Multi-Type Anomaly Detection}
\titlerunning{Explainable Deep Convolutional Multi-Type Anomaly Detection}
%
\author{Alex George\orcidID{0009-0008-5799-4161} \and
Lyudmila Mihaylova\orcidID{0000-0001-5856-2223} \and
Sean Anderson\orcidID{0000-0002-7452-5681}} 
\authorrunning{A. George et al.}
%
\institute{School of Electrical and Electronic Engineering, University of Sheffield, UK\\
\email{\{ageorge4,l.s.mihaylova,s.anderson\}@sheffield.ac.uk}}
\maketitle              

\begin{abstract}
Explainable anomaly detection methods often have the capability to identify and spatially localise anomalies within an image but lack the capability to differentiate the \emph{type} of anomaly. Furthermore, they often require the costly training and maintenance of separate models for each object category. The lack of specificity is a significant research gap because identifying the type of anomaly (e.g., ``Crack'' vs. ``Scratch'') is crucial for accurate diagnosis that facilitates cost-saving operational decisions across diverse application domains. While some recent large-scale Vision-Language Models (VLMs) have begun to address this, they are computationally intensive and memory-heavy, restricting their use in real-time or embedded systems.
We propose MultiTypeFCDD, a simple and lightweight convolutional framework designed as a practical alternative for explainable multi-type anomaly detection. MultiTypeFCDD uses only image-level labels to learn and produce multi-channel heatmaps, where each channel is trained to correspond to a specific anomaly type. The model functions as a single, unified framework capable of differentiating anomaly types across multiple object categories, eliminating the need to train and manage separate models for each object category.
We evaluated our proposed method on the Real-IAD dataset and it delivers competitive results (96.4\% I-AUROC) at just over 1\% the size of state-of-the-art VLM models used for similar tasks. This makes it a highly practical and viable solution for real-world applications where computational resources are tightly constrained. 

\end{abstract}

\section{Introduction}









Anomaly detection is a critical aspect in computer vision where the core objective is to identify patterns in visual data that do not conform to expected normal behaviour \cite{pang_2021_deep_anomaly_detection_survey}. Recent years have witnessed significant adoption of deep learning-based anomaly detection in domains ranging from medical imaging \cite{fernando2021deep}, autonomous driving \cite{bogdoll_2022_anomaly_driving_survey} to industrial manufacturing \cite{alzarooni2025anomaly}. An important aspect of these sophisticated models is their explainability  -- by explainability in this paper, we mean spatially localising the anomaly within the input image. Being able to understand why the model flagged something as an anomaly not only builds trust, but also helps the end users to make better and informed decisions, especially in sensitive areas like healthcare and autonomous driving \cite{arrieta2020explainable,li2023survey}. By definition, anomalies are rare events, leading to a class imbalance where the normal samples vastly outnumber the abnormal ones \cite{ruff_2021_anomaly_review}. Consequently, many state-of-the-art methods approach this problem of explainable anomaly detection from a one-class classification perspective \cite{defard2021padim,deng2022anomaly,liu2023simplenet,liznerski_explainable_2021,roth2022towards}.

\begin{figure}[h]
\includegraphics[width=\textwidth]{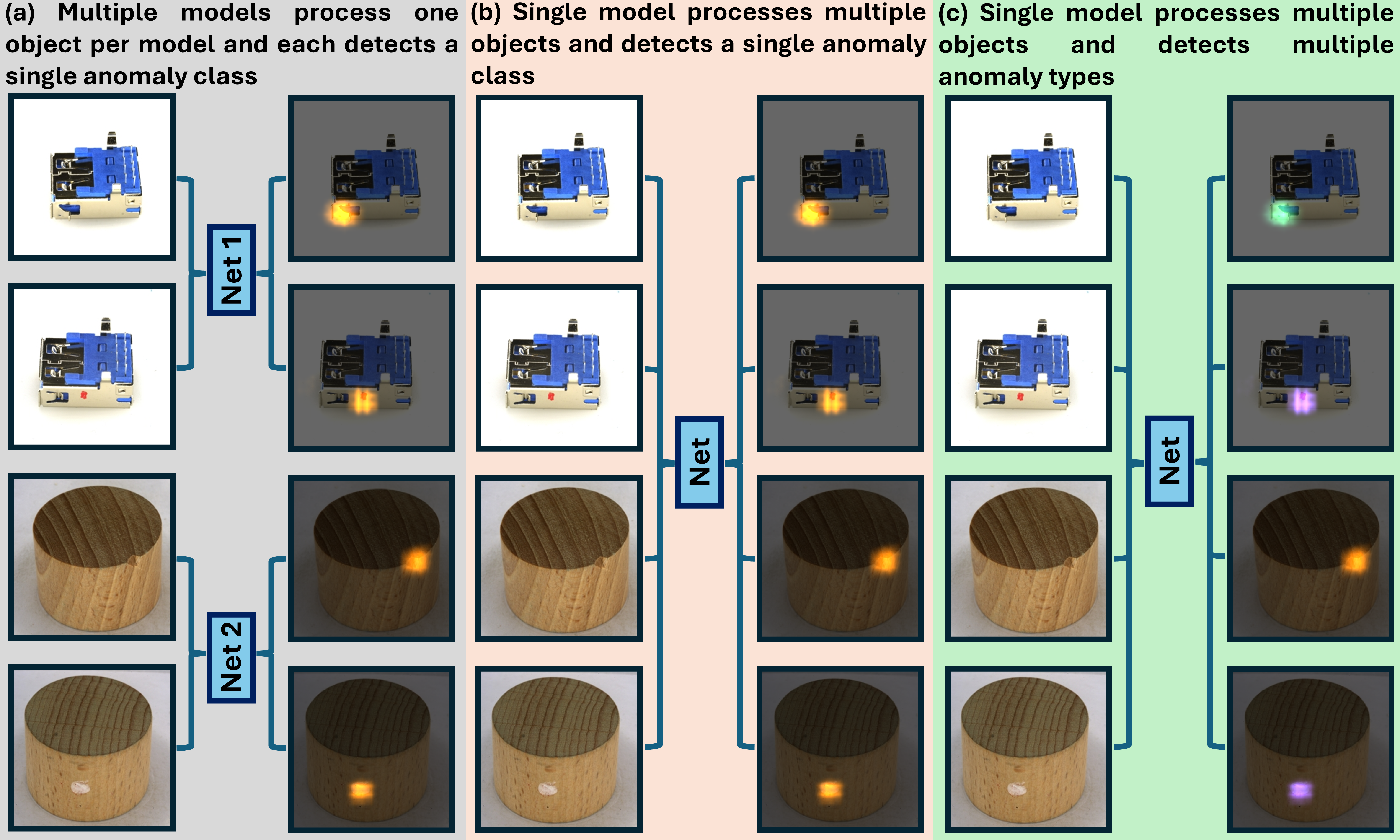}
\caption{Illustration of multi-type anomaly detection on two objects (USB and Woodstick) from the Real-IAD dataset \cite{wang2024real}. (a) In multi-class anomaly detection, most explainable methods require training on separate networks to detect anomalies in each object class. (b) Recent research has introduced unified frameworks to detect anomalies across multiple objects. However, both (a) and (b) are only capable of detecting an anomaly and not what type of anomaly. (c) Our proposed framework enables the detection of multiple types of anomalies occurring across diverse objects using a single, unified model.} \label{multitypeAD_figure}
\end{figure}

A key limitation of existing anomaly detection methods is that they often require training separate models for each object class, which limits scalability and practicality in real-world applications, as illustrated in Fig.~\ref{multitypeAD_figure}(a). Even “multi-class anomaly detection” approaches address this only partially \cite{you2022unified}, as unified models can detect anomalies but do not distinguish between different types of anomaly, as shown in Fig.~\ref{multitypeAD_figure}(b). This inhibits real-world decision making where understanding the type of anomaly is critical for determining the appropriate course of follow-up action \cite{foorthuis2021nature}. 

Recently, methods have begun to address this problem by exploiting vision-language models (VLMs) \cite{mokhtar2025detect,sadikaj2025multiads}, combining text prompts with visual features to segment multiple defect types. However, these models are computationally intensive, leading to substantial energy consumption and which translates into significant carbon emissions and financial costs \cite{papa2024survey} 
Additionally, such architectures are often not well-suited to systems with tight constraints on computational power and latency \cite{marshall2025transformers}, such as embedded systems, robotics and industrial automation. 
This highlights a clear need to explore alternative approaches that prioritise computational efficiency and practical deployability. 

To address this research gap, our work proposes a novel deep convolutional multi-type anomaly detection method. We base the approach on the Fully Convolutional Data Description (FCDD) \cite{liznerski_explainable_2021}.
%
This enables the simultaneous detection and spatial localisation within an image for multiple pre-defined anomaly types. 
Our research makes three main contributions:
\begin{itemize}
    \item We contribute a novel fully convolutional model for multi-type anomaly detection with multiple output channels sensitised to different types of anomaly, which we refer to as MultiTypeFCDD.
    \item We contribute a novel loss function and training framework for multi-type anomaly detection, extending the original FCDD framework.
    \item We demonstrate, on the publicly available Real-IAD dataset \cite{wang2024real}, that our proposed lightweight framework retains competitive performance to state-of-the-art models while operating with significantly fewer parameters and lower inference times.
\end{itemize}

The paper is structured as follows. In Section 2, we review related work to explainable anomaly detection. In Section 3, we describe the proposed method for multi-type anomaly detection. In Section 4, we describe the experimental setup to evaluate the proposed method, including the publicly available RealIAD dataset and benchmark algorithms.  In Section 5, we provide the results and in Section 6, we conclude the work. 


\section{Related Works}

\subsection{Deep Learning for Anomaly Detection}

Early approaches for anomaly detection utilised classical machine learning methods such as One-Class Support Vector Machines (OC-SVM) \cite{scholkopf2001estimating}, Support Vector Data Descriptions (SVDD) \cite{tax2004support} and Isolation Forests \cite{liu2008isolation}. Methods such as SVDD have been extended with deep learning methods in Deep SVDD \cite{ruff2018deep} to improve the processing of complex data sets, which has improved performance in cases where it is challenging to obtain a useful feature representation \cite{pang_2021_deep_anomaly_detection_survey}. 


\subsection{Explainable Anomaly Detection and Segmentation}

The black-box nature of deep learning models has driven the need to develop methods to aid transparency in the decision-making process \cite{arrieta2020explainable}. 
Several techniques have been specifically designed for explainable anomaly detection, including feature-embedding methods such as PaDiM \cite{defard2021padim}, PatchCore \cite{roth2022towards} and SoftPatch \cite{jiang2022softpatch}, and end-to-end trainable methods such as FCDD \cite{liznerski_explainable_2021}, SimpleNet \cite{liu2023simplenet} and Reverse Distillation (RD) \cite{deng2022anomaly}.

As the above approaches rely on image-level labels (or no labels) rather than pixel-level annotations for training, they can be regarded as closely related to weakly supervised semantic segmentation (WSSS) \cite{jiang2023weakly,yang2024anomaly}. In WSSS, models produce dense pixel-wise maps from weak annotations such as image-level labels or bounding boxes. Consequently, anomaly detection and segmentation under weak supervision naturally aligns within this paradigm.

\subsection{Multi-Class vs Multi-Type Anomaly Detection}

The methods discussed above frame anomaly detection from a one-class classification perspective, where the goal is to learn the representation of the normal data, and thereby separate everything anomalous as a single, separate class.  This approach is sufficient for many anomaly detection benchmarks such as MVTec-AD \cite{bergmann2021mvtec} and VisA \cite{zou2022spot}. 
%

Recent literature has started to explore and use the terminology ``multi-class anomaly detection'' \cite{he2024mambaad,you2022unified}, which typically refers to identifying a single anomaly class across multiple objects. Additionally, most existing models are designed to segment anomalies within a single object, 
meaning that separate models might be required to process each object. 

Efforts have been made to create unified models for multi-class anomaly detection, 
such as UniAD \cite{you2022unified}. However, these models cannot distinguish between different types of anomaly. In short, multi-type anomaly detection remains an area that is still under-explored \cite{mokhtar2025detect}. 


\subsection{Vision-Language Models for Anomaly Characterisation}

Recent research has made use of vision-language models (VLMs) \cite{chen2025can,gu2024anomalygpt,jeong2023winclip} to tackle multi-type anomaly detection. These models combine text prompts with image features, allowing them to both classify anomalies as well as create pixel-wise anomaly maps highlighting anomalous regions in the image. Two methods in particular, MultiADS \cite{sadikaj2025multiads} and VELM \cite{mokhtar2025detect}, have demonstrated this approach for industrial anomaly detection.

Despite their advantages, VLMs are generally heavy, computationally intensive and have higher inference time, making them less well suited to real-time or embedded systems \cite{garcia2019estimation,papa2024survey,strubell2020energy}. 


\subsection{Research Gap and Our Contribution}

The literature contains two main types of deep anomaly detectors: in one group, there are  lightweight anomaly detection models, based on convolutional networks, which are typically limited to detecting only one type of anomaly.  In the other group, we have powerful VLMs which can perform multi-type anomaly detection and segmentation but are computationally expensive. This identifies a clear research gap: the need for a unified, lightweight model with inherent explainability, capable of multi-type anomaly detection and segmentation.

We introduce an extension of the FCDD [28] framework, which we term MultiTypeFCDD, that generates multi-channel anomaly heatmaps, with each channel representing a specific type of anomaly. This design enables the simultaneous detection and localization of multiple predefined anomaly types.

\section{Proposed Method}

\subsection{Model}

\begin{figure}[ht]
\includegraphics[width=\textwidth]{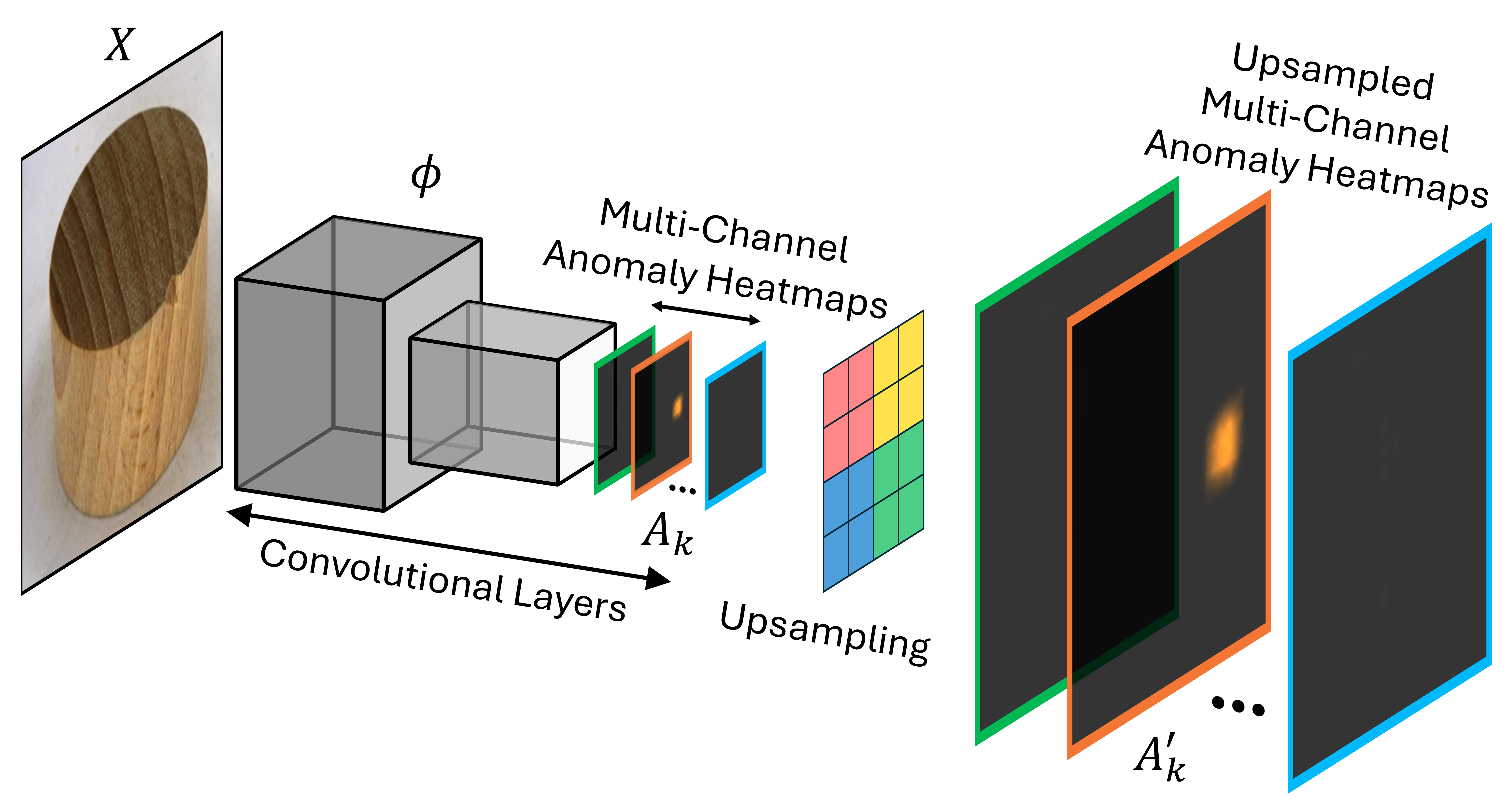}
\caption{Illustration of MultiTypeFCDD framework. An input image $X$ is fed into a convolutional network $\phi$ which produces multi-channel outputs $A_k$ corresponding to different anomaly types. These outputs are then resized to heatmaps $A'_k$ that match the input image dimensions.} \label{multichannel_FCDD}
\end{figure}

Our method builds upon the fully convolutional architecture in FCDD \cite{liznerski_explainable_2021} which produces explainable anomaly heatmaps. The key novelty lies in the output layer. Unlike the original FCDD architecture which generates a single anomaly heatmap, we extend the architecture such that the output layer produces multi-channel anomaly heatmaps. Specifically, we replace the final layer with a $1\times1$ convolutional layer with depth equal to the number of types of anomaly, enabling the model to produce a separate heatmap for each anomaly class. Figure \ref{multichannel_FCDD} shows the outline of our proposed framework.

Let $X_i \in \mathbb{R}^{h \times w \times c}, i \in \{1,\cdots, N\}$ be the set of input images. Let $M$ be the number of anomaly classes and $y_{ik} \in \{0,1\}$ be the binary labels for each anomaly class $k \in \{1,\cdots, M\}$ assigned to the image $X_i$. This indicates the presence $(1)$ or absence $(0)$ of the anomaly class. An image $X_i$ is defined as Normal if all anomaly classes are absent, i.e., $\forall k, y_{ik}=0$. 
Let $\phi(X_i;W):\mathbb{R}^{h \times w \times c} \rightarrow \mathbb{R}^{u \times v \times M}$ be the fully convolutional network with weights $W$. A pseudo-Huber loss function is applied to the output of $\phi$ and generates a low-resolution anomaly heatmap $A_k(X_i) \in \mathbb{R}^{u \times v}$ for each class $k$ defined as

\begin{equation}
A_k\left(X_i\right)=\sqrt{{\phi\left(X_i;W\right)}^2+1}-1.
\label{eq:fcdd_anomalyscore_eqn}
\end{equation}

The overall anomaly score $z_{ik}$ for any input image is computed as the L1-norm of the corresponding anomaly heatmap $A_k\left(X_i\right)$, defined as
\begin{equation}
z_{ik} = \frac{1}{u\cdot v}\norm{A_k\left(X_i\right)}_{1},
\label{eq:mean_anomaly_score}
\end{equation}
where $u$ and $v$ correspond to the dimensions of $A_k$ which is the same for each heatmap. 
The heatmaps generated by the model are of lower resolution than the input due to downsampling through convolution and pooling layers. Therefore, we upsample each heatmap using bilinear interpolation to generate $A'_k$ that matches the input image dimensions, thereby producing a one-to-one correspondence to each pixel in the anomaly heatmap to the input image.

\subsection{Loss Function}

The standard FCDD loss function is designed for binary anomaly detection where each image is either normal or anomalous. In our multi-type anomaly detection setting, we extend the FCDD loss function to accommodate each type of anomaly. 
The model must be trained under two conditions: 1. there must examples of anomalies available (so it cannot be trained as a one-class classifier unlike the original FCDD) and 2. the training data set must contain images with single anomaly types (although at inference time the model can predict multiple types of anomaly in a single image).

The proposed objective function is defined as
\begin{equation}
\min_{W} \ \frac{1}{MN} \sum_{i=1}^{N} \sum_{k=1}^{M}
(1 - y_{ik})\, z_{ik} - y_{ik} \log\!\big(1 - e^{\displaystyle -z_{ik}}\big).
\label{eq:multitype_fcdd_loss_eqn}
\end{equation}

Using only image-level labels, the model learns to identify the discriminative regions in the images containing the target class and those without it. The loss function encourages the model to localise anomalies by maximising pixel-level anomaly scores in spatial regions where anomalies are present while suppressing the scores elsewhere, thereby enabling the generation of class-specific anomaly heatmaps in a weakly supervised manner.

\section{Experimental Setting}

\subsection{Dataset}

Many existing anomaly detection datasets like MVTec-AD \cite{bergmann2021mvtec} and VisA \cite{zou2022spot} focus on object-specific defects, where the anomalies are closely tied to the specific object. This makes them unsuitable for the task of multi-type anomaly detection, where our approach is designed to detect and differentiate various types of defects found across various objects.

To address this limitation, we adopt the Real-IAD dataset \cite{wang2024real}, which provides image-level and pixel-level annotations for various defect types commonly observed across a wide variety of objects. The Real-IAD dataset comprises 151,050 images of 30 objects, consisting of 99,721 normal images and 51,329 anomalous images, spanning 8 defect categories. Table \ref{tab:defect_code_table} summarises the various defect codes used in the dataset.

\begin{table}[ht]
    \centering
    \caption{Defect codes and their definitions in the Real-IAD dataset.}
    \begin{tabular}{cc@{\hspace{2em}}cc}
    \cmidrule(r{1.8em}){1-2} \cmidrule(l{0.0em}){3-4}
    \textbf{Code} & \textbf{Defect Type} & \textbf{Code} & \textbf{Defect Type} \\
    \cmidrule(r{1.8em}){1-2} \cmidrule(l{0.0em}){3-4}
    AK & Pit & PS & Damage\\
    BX & Deformation & QS & Missing Parts\\
    CH & Abrasion & YW & Foreign Objects\\
    HS & Scratch & ZW & Contamination\\
    \cmidrule(r{1.8em}){1-2} \cmidrule(l{0.0em}){3-4}
    \end{tabular}
    \label{tab:defect_code_table}
\end{table}

We follow the same train/test split procedure as in the original paper \cite{wang2024real} and use three training sets with noise ratios 
\begin{equation}
    \alpha \in \{0.1, 0.2, 0.4 \} 
\end{equation}
which is a parameter that controls the number of anomalous samples mixed into the training data.

\subsection{Balanced Dataset Sampling: Redefining Epochs}

Anomaly detection tasks inherently suffer from data imbalance, as datasets contain fewer anomalous samples than normal images. Given that we require images of various types of anomalies rather than ``any'' anomaly, the data imbalance problem becomes even more severe. Training a model on an imbalanced dataset can lead to overfitting to patterns in normal images and reduced sensitivity to anomalies \cite{ruff_2021_anomaly_review}. To address this, we adopt a balanced sampling strategy during training, inspired by the sampling procedure for binary anomaly detection in \cite{liznerski_explainable_2021}, where images from normal and each anomaly class are uniformly sampled per iteration.

While this improves performance and helps the model better distinguish between normal and various anomaly types, it has an important computational implication: the number of iterations required to complete an epoch increases. Unlike standard training epochs that process each image exactly once, balancing requires the reshuffling and resampling under-represented classes, necessitating the analytical computation of the total number of iterations per epoch.




The problem is analogous to the generalised coupon collector problem \cite{shank2013coupon}, where each ``coupon'' corresponds to an image from a given target class. Let each class $i \in \{1,2, \cdots,n\}$ have a target quota of $m_i$ (number of images per class) and is randomly chosen with equal probability. During sampling, if any class $i$ has reached its full quota $m_i$, its images are reshuffled and can be sampled again in subsequent iterations. Let $E[T]$ be the expected number of iterations needed to sample at least $m_i$ images from every class. Although a closed-form expression for $E[T]$ is given in \cite{shank2013coupon}, direct computation is practically infeasible, especially when $m_i$ is large $(m_i \gg 1)$ as it involves computing factorials of $m_i$. Therefore, in practice, one can use Monte Carlo simulation to estimate $E[T]$ empirically as

\begin{equation}
E[T] \approx \mu_T = \frac{1}{N_\text{trials}} \sum_{j=1}^{N_\text{trials}} T_j,
\end{equation}
where $T_j$ is the number of iterations in trial $j$, and $N_{trials}$ is the total number of Monte Carlo trials. This gives the empirical mean iterations per epoch, $\mu_T$. During training, images are processed in mini-batches. Therefore, the mean number of mini-batch iterations per epoch, which we refer to as balanced epochs, can be computed as $\mu^{batch}_T$.

Table \ref{tab:montecarlo_results} shows Monte Carlo simulation results (10000 trials) on the three different Real-IAD training datasets across 9 image classes (8 anomaly classes and 1 normal class) at a mini-batch size of 32. When interpreted under the standard epoch definition (where each image is sampled exactly once), a balanced epoch corresponds to several standard epochs, e.g. at $\alpha = 0.1$, one balanced epoch equals 8.41 standard epochs. Therefore, even when the model appears to converge in only a few balanced epochs, it has actually seen the data multiple times, achieving better class coverage and more stable learning.

\begin{table}[ht]
\centering
\setlength{\tabcolsep}{3pt}
\caption{Monte Carlo simulation results to estimate iterations per balanced epoch.}
\begin{tabular}{lccccc}
\toprule
\textbf{$\alpha$} & \textbf{Total Images} & \textbf{Image Classes} & $\mu^{batch}_T$ & \textbf{Std. Iterations} & \textbf{Std. Epochs} \\
\midrule
0.1 & 57,840 & 9 & 15,206.14 & 1807.5 & 8.41 \\
0.2 & 57,840 & 9 & 14,128.68 & 1807.5 & 7.82 \\
0.4 & 57,840 & 9 & 11,960.70 & 1807.5 & 6.62 \\
\bottomrule
\end{tabular}
\label{tab:montecarlo_results}
\end{table}

\subsection{Training Setup}

\subsubsection{Model Architecture:} The backbone of the network employs the first three stages of an Inception-ResNet-v2 encoder \cite{szegedy2017inception} pre-trained on ImageNet \cite{deng2009imagenet}, with all layers frozen to retain generic feature representations. The encoder is appended with a convolutional head which contains three sequential blocks: (1) two convolutional blocks ($3 \times 3$ convolution with 512 filters $+$ batch normalisation $+$ ReLU), (2) a $1 \times 1$ convolution layer with the number of output channels equal to the number of anomaly types, and (3) a differentiable function layer applying pseudo-Huber loss for each anomaly channel as given in Eq. \ref{eq:fcdd_anomalyscore_eqn}. 

\subsubsection{Hyperparameter Tuning:} For robust training, data augmentations were applied randomly to $50\%$ of the images in the training dataset. These included random rotations within $\pm15^\circ$, random translations within $\pm20$ pixels along both X and Y directions, and random brightness and contrast adjustments. The network was trained using the Adam optimiser with a mini-batch size of 32 with an initial learning rate of 0.0001.

\subsubsection{Hardware Configuration:} All experiments were conducted on a desktop PC equipped with an NVIDIA RTX 4070 GPU and an AMD Ryzen 5 5500 CPU.

\subsection{Metrics}

We follow the Real-IAD \cite{wang2024real} benchmark protocol, where model performance is evaluated at both the image and pixel levels in terms of the Receiver Operating Characteristic (ROC) curve, defined by the True Positive Rate (TPR) versus the False  Positive Rate (FPR). 
We summarise the ROC curve using the area under the ROC curve (AUROC) for both the Image-level AUROC (I-AUROC) and Pixel-level AUROC (P-AUROC). For image-level metrics, we compute the anomaly score as the mean value of the anomaly heatmap per channel as given in Eq. \ref{eq:mean_anomaly_score}.


To assess the performance of spatial localisation of the anomaly, we calculate the per-region overlap (PRO) score  \cite{bergmann2021mvtec}. Let $C_{i,k}$ represent the set of pixels for the $k^{th}$ ground-truth component in image $i$, and $P_i(\tau)$ be the set of predicted anomalous pixels for image $i$ at a given threshold $\tau$; the PRO score is defined as the average coverage of these components across the entire dataset,
\begin{equation}
    \text{PRO}(\tau) = \frac{1}{N} \sum_{i} \sum_{k} \frac{|P_i(\tau) \cap C_{i,k}|}{|C_{i,k}|}
\end{equation}
where $N$ denotes the total count of ground-truth components in the dataset.  We use the Area Under the PRO curve (AUPRO) to summarise the PRO score in a single metric, which is defined as the normalised area under the PRO-FPR curve up to an FPR limit of 0.3.


\subsection{Benchmarking}

To benchmark MultiTypeFCDD, we compare it to several other methods that perform explainable anomaly detection (using the Real-IAD dataset), including PaDiM \cite{defard2021padim}, CFlow \cite{gudovskiy2022cflow}, PatchCore \cite{roth2022towards}, SimpleNet \cite{liu2023simplenet}, DeSTSeg \cite{zhang2023destseg}, RD \cite{deng2022anomaly}, and UniAD \cite{you2022unified}. These benchmark methods do not distinguish types of anomaly, so we only compared performance by treating all anomaly types as a single class. Although MultiTypeFCDD uses weak image-level labels during training, these do not provide spatial supervision, so comparison with unsupervised one-class methods remains fair. To handle multiple objects, we compared multiple models trained on separate objects (where necessary). We compared performance in terms of I-AUROC to characterise anomaly detection at the image level, and AUPRO to characterise the success of spatial localisation of the anomaly.

\section{Results}

\subsection{Multi-Type Anomaly Detection}

The results of multi-type anomaly detection using our method are presented in Table \ref{tab:multi-type_detection_performance}. At the image-level, mean I-AUROC reports over 95\% across varying $\alpha$ values, indicating the model's capability to discriminate between normal and anomaly-specific images. At the pixel level, the model maintains robust localisation performance with a mean P-AUROC above 93\% and steady AUPRO improvements as $\alpha$ the number of iterations increases. In general, the model has a strong capability of detecting various types of anomalies. However, certain anomaly types, such as Foreign Objects (YW), exhibit lower performance because they have fewer samples, fewer ground-truth pixels and less consistent patterns due to their high semantic diversity when compared to classes like Deformation (BX). 

\begin{figure}[h!]
\includegraphics[width=\textwidth]{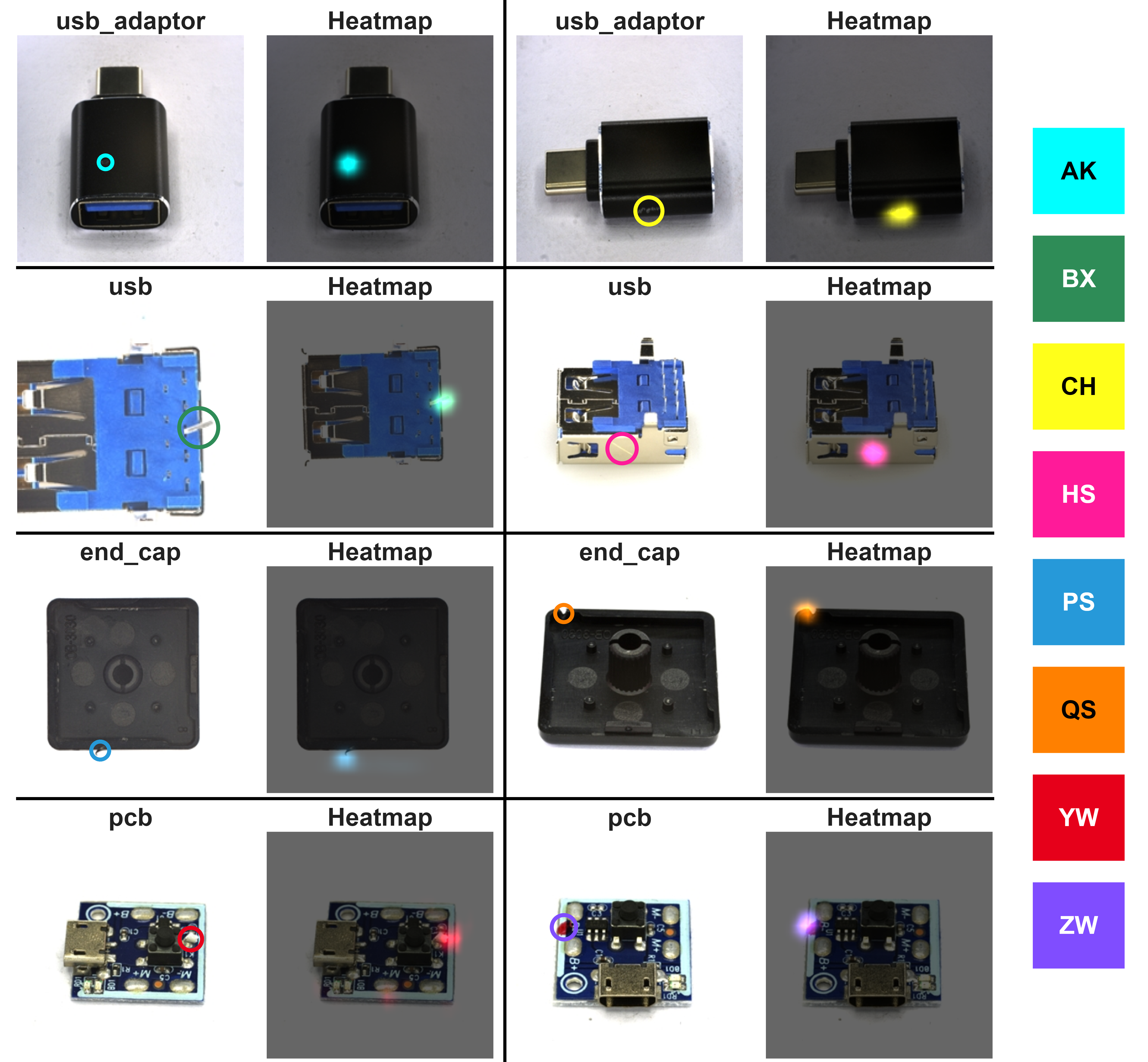}
\caption{Examples of multi-type anomaly detection across multiple objects ($\alpha = 0.2$).  Each pair of images in the grid shows the input image with the ground-truth anomaly highlighted by the circle (left) and the corresponding predicted anomaly heatmap by MultiTypeFCDD overlaid on the input image (right), with colours defining each anomaly type indicated in the legend on the right.} \label{defect_heatmap_per_class}
\end{figure}

\begin{figure}[h!]
\centering
\includegraphics[width=0.9\textwidth]{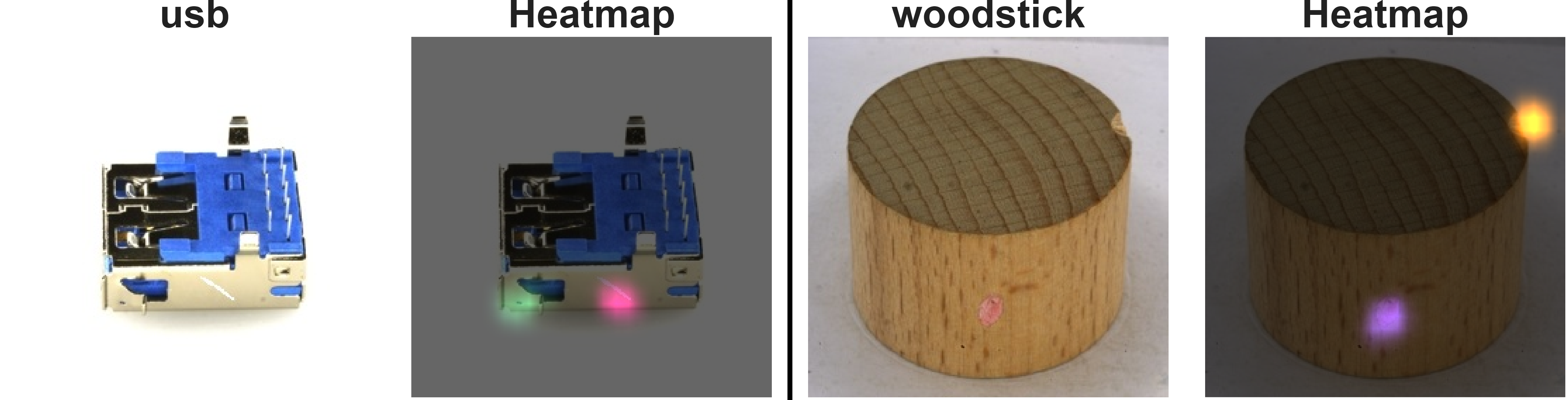}
\caption{Example of multi-type anomaly detection using manually edited synthetic test images containing multiple co-occurring anomalies ($\alpha = 0.2$). For each image, the corresponding overlaid anomaly heatmap highlights all detected anomalous regions.} \label{defect_heatmap_custom_images}
\end{figure}

\begin{figure}[h!]
\includegraphics[width=\textwidth]{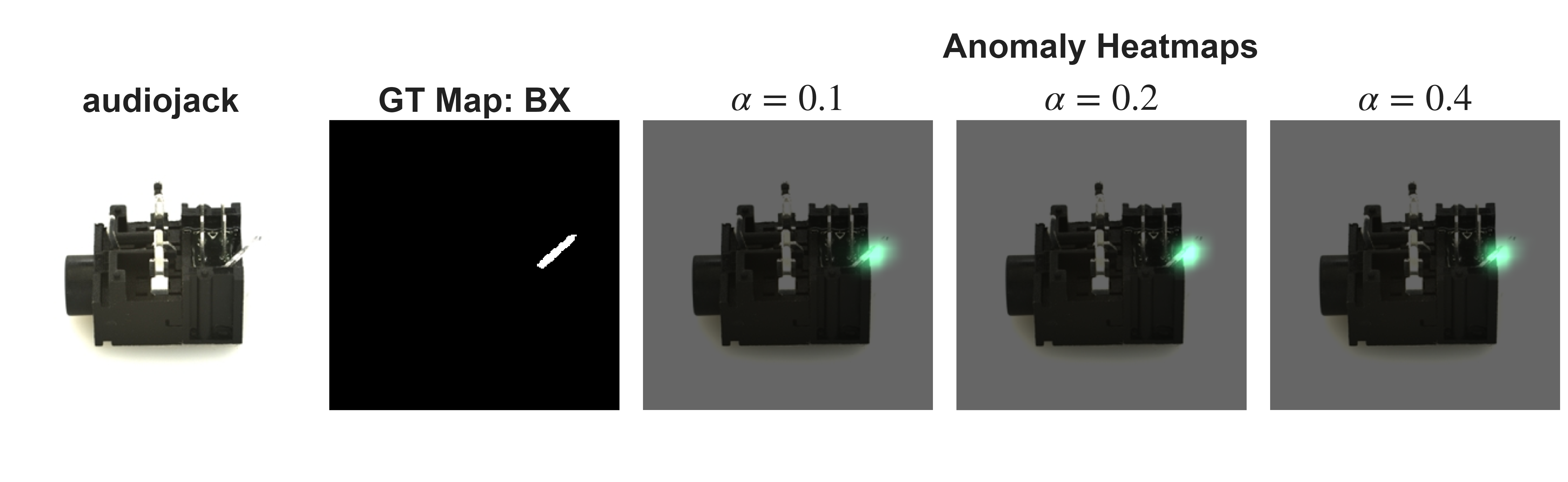}
\caption{Comparison of anomaly heatmaps for a test object (audiojack) across different $\alpha$ levels: 0.1, 0.2, 0.4.} \label{heatmap_comparison}
\end{figure}

Figure \ref{defect_heatmap_per_class} gives a visualisation of anomaly localisation across the eight defect classes for various objects at $\alpha = 0.2$. Because the Real-IAD dataset contains only one defect type per image, Figure \ref{defect_heatmap_custom_images} includes custom-generated examples created by merging two defect instances of the same object class into a single sample. This example demonstrates that the model is capable of identifying and localising multiple anomalies within one image. Furthermore, to illustrate the localisation behaviour across multiple $\alpha$ values, Fig. \ref{heatmap_comparison} provides a visual comparison of anomaly heatmaps for a representative test object.

\begin{table}[ht]
\centering
\caption{Image and pixel-level multi-type anomaly detection performance across different $\alpha$ levels: 0.1, 0.2, 0.4.}
\setlength{\tabcolsep}{2.5pt} 
\begin{tabular}{lccccccccc}
\toprule
 \multirow{2}{*}{\textbf{Anomaly Type} \vspace{-0.15cm}} & \multicolumn{3}{c}{\textbf{I-AUROC}} & \multicolumn{3}{c}{\textbf{P-AUROC}} & \multicolumn{3}{c}{\textbf{AUPRO}} \\
\cmidrule(lr){2-4} \cmidrule(lr){5-7} \cmidrule(lr){8-10}
& \textbf{0.1} & \textbf{0.2} & \textbf{0.4} & \textbf{0.1} & \textbf{0.2} & \textbf{0.4}
& \textbf{0.1} & \textbf{0.2} & \textbf{0.4}  \\
\midrule
Pit (AK)             & 95.2& 95.7 & 96.2 & 97.2& 94.8 & 97.0 & 84.8& 80.7& 85.6\\
Deformation (BX)     & 99.6& 99.3 & 99.4 & 98.4& 96.3 & 96.1 & 87.4& 86.4& 87.4\\
Abrasion (CH)        & 98.3& 98.2 & 99.1 & 96.0& 95.5 & 96.4 & 81.8& 84.0& 83.7\\
Scratch (HS)         & 90.0& 94.4 & 95.9 & 85.3& 90.5 & 88.9 & 64.0& 72.5& 67.6\\
Damage (PS)          & 98.7& 98.6 & 98.8 & 96.0& 96.4 & 97.8 & 79.2& 88.3& 88.7\\
Missing Parts (QS)   & 92.0& 94.3 & 96.1 & 84.2& 91.5 & 91.8 & 70.6& 82.3& 80.5\\
Foreign Objects (YW) & 98.7& 99.3 & 99.4 & 95.0& 93.3 & 91.5 & 57.5& 60.3& 49.2\\
Contamination (ZW)   & 92.9& 93.2 & 95.0 & 95.6& 95.8 & 96.6 & 79.3& 81.9& 82.3\\
\midrule
\textbf{Mean}   & \textbf{95.7} & \textbf{96.6} & \textbf{97.5} & \textbf{93.5} & \textbf{94.3} & \textbf{94.5} & \textbf{75.6}& \textbf{79.6}& \textbf{78.1}\\
\bottomrule
\end{tabular}
\label{tab:multi-type_detection_performance}
\end{table}


\subsection{Multi-Class Anomaly Detection}

We also evaluate MultiTypeFCDD in a multi-class anomaly detection setting using a single-channel anomaly output. This allows for fair benchmarking against state-of-the-art multi-class anomaly detection methods, as summarised in Table~\ref{tab:multi_class_performance}. While unsupervised methods show degradation in performance as the proportion of anomalous samples in the training dataset increases, weakly supervised approaches like ours, even with just image-level labels, benefit from additional anomalous samples and show improved performance. MultiTypeFCDD achieves consistently high I-AUROC (above 94\%) across different $\alpha$ levels, demonstrating strong anomaly detection capability and generalisation across object categories. Although its AUPRO scores, which indicate anomaly localisation performance, are lower than those of methods such as PatchCore or SoftPatch, this is mainly because these approaches train separate models for each object class, allowing more precise localisation, and the results are averaged. In contrast, unified frameworks like MultiTypeFCDD and UniAD utilise a single model trained across all object classes, which makes fine-grained localisation more challenging but substantially reduces the total number of parameters, requiring only a fraction of the memory and computational overhead while maintaining competitive overall performance.

\begin{table}[ht]
\centering
\setlength{\tabcolsep}{2.5pt} 
\caption{Multi-Class Anomaly Detection Performance for $\alpha$-values 0.1, 0.2, 0.4. The asterisk \textbf\textsuperscript{*} indicates that the method uses a single model across all object classes.}
\begin{tabular}{lcccccccc}
\toprule
\multirow{2}{*}{\textbf{Method} \vspace{-0.15cm}} &
\multicolumn{3}{c}{\textbf{I-AUROC}} &
\multicolumn{3}{c}{\textbf{AUPRO}} &
\multirow{2}{*}{\textbf{Parameters} \vspace{0.3cm} } \\
\cmidrule(lr){2-4} \cmidrule(lr){5-7}
& \textbf{0.1} & \textbf{0.2} & \textbf{0.4}
& \textbf{0.1} & \textbf{0.2} & \textbf{0.4}
& \smash{\raisebox{0.5ex}{\textbf{(M)}}} \\
\midrule
PaDiM~\cite{defard2021padim} & 81.9 & 80.1 & 77.0 & 86.4 & 86.5 & 86.1 & 68.9\\
CFlow~\cite{gudovskiy2022cflow} & 83.9 & 79.6 & 78.0 & 90.7 & 90.7 & 90.2 & 30.4\\
PatchCore~\cite{roth2022towards} & 90.4 & 89.5 & 88.1 & \textbf{94.2} & \textbf{93.9} & 92.4 & 68.9\\
SoftPatch~\cite{jiang2022softpatch} & 90.9 & 90.5 & 89.3 & 92.9 & 92.9 & \textbf{92.5} & 68.9\\
SimpleNet~\cite{liu2023simplenet} & 79.9 & 75.9 & 74.7 & 83.9 & 83.2 & 70.5 & 68.9\\
DeSTSeg~\cite{zhang2023destseg} & 85.6 & 80.3 & 74.4 & 86.9 & 83.2 & 75.5 & 37.9\\
RD~\cite{deng2022anomaly} & \textbf{95.1} & 94.9 & 94.7 & 87.6 & 83.2 & 81.5 & 68.9\\
UniAD\textbf\textsuperscript{*}~\cite{you2022unified} & 84.2 & 82.8 & 80.1 & 87.7 & 87.3 & 86.6 & 7.6\\
\midrule
\textbf{MultiTypeFCDD}\textbf\textsuperscript{*} & 94.1 & \textbf{95.5} & \textbf{96.4} & 73.3 & 73.7 & 78.1 & \textbf{5.5}\\
\bottomrule
\end{tabular}
\label{tab:multi_class_performance}
\end{table}

\subsection{Computational Performance}


To emphasise the practical deployability of our approach, we benchmark its computational footprint against MultiADS \cite{sadikaj2025multiads} presented in Table \ref{tab:comp_perf}. While MultiADS is a potential baseline in multi-type anomaly detection, it employs a VLM trained with image-text prompts, which differs substantially from our approach that uses only images and standard convolutional layers. Consequently, we do not perform a direct comparison under the multi-type anomaly detection setting as it would not meaningfully reflect the strengths of our approach. 

Remarkably, MultiTypeFCDD operates with only 5.5 M parameters, marking a significant 98.7\% reduction compared to the 427.9 M parameters used in MultiADS. This massive reduction in parametric load directly translates to significantly faster inference times, with ours requiring only 41.7 ms and 1.4ms on CPU and GPU respectively, compared to 1824.3 ms (CPU) and 33.7 ms (GPU) for MultiADS. This comparison demonstrates that complex VLMs are not strictly necessary to achieve reliable multi-type anomaly detection. These findings align with prior observations where large-scale VLMs are noted to be computationally and memory-intensive, restricting their usability in real-time or embedded systems~\cite{garcia2019estimation,marshall2025transformers,strubell2020energy}. In contrast, our lightweight architecture achieves competitive performance while substantially reducing memory and inference costs.


\begin{table}[h!]
\centering
\setlength{\tabcolsep}{4pt}
\caption{Computational performance comparison.}
\label{tab:comp_perf}
\begin{tabular}{lccc}
\toprule
\multirow{2}{*}{\textbf{Model} \vspace{-0.15cm}}
& \multirow{2}{*}{\textbf{Parameters} \vspace{0.3cm} }
& \multicolumn{2}{c}{\textbf{Inference Time (ms/img)}} \\
\cmidrule(lr){3-4}
 & \smash{\raisebox{0.5ex}{\textbf{(M)}}} 
 & \multicolumn{1}{c}{\scriptsize\textbf{\ \ \ \ \ \ CPU}} 
 & \multicolumn{1}{c}{\scriptsize\textbf{\ \ \ GPU}} \\
\midrule
MultiADS \cite{sadikaj2025multiads} & 427.9 & \ \ \ \ \ 1824.3 & \ \ \ 33.7 \\
\textbf{MultiTypeFCDD} & \textbf{5.5} & \ \ \ \ \ \textbf{41.7} & \ \ \ \textbf{1.4} \\
\bottomrule
\end{tabular}
\end{table}

\subsection{Ablation Study}

Table \ref{tab:ablation_study} presents an ablation study on MultiTypeFCDD analysing the impact of backbone, downsampling stage, convolutional blocks, and balanced training on model performance and inference times. We report the mean multi-type anomaly detection metrics across all anomaly classes using the training dataset with $\alpha = 0.1$. 

\begin{table}[h!]
\centering
\caption{Ablation study on model performance. We analyse the effects of (a) Backbone, (b) Downsampling Stage, (c) Convolutional Blocks, and (d) Balanced Training Strategy. Default settings are indicated by a checkmark \checkmark.}
\label{tab:ablation_study}
\setlength{\tabcolsep}{1.5pt} 
\begin{tabular}{lcccccc} 
\toprule
\multirow{2}{*}{\textbf{Option} \vspace{-0.15cm}}
& \multirow{2}{*}{\textbf{Params} \vspace{0.3cm} }
& \multirow{2}{*}{\footnotesize{\textbf{I-AUROC}} \vspace{-0.15cm} }
& \multirow{2}{*}{\footnotesize{\textbf{P-AUROC}} \vspace{-0.15cm} }
& \multirow{2}{*}{\footnotesize{\textbf{AUPRO}} \vspace{-0.15cm} }
& \multicolumn{2}{c}{\textbf{Inf. Time (ms/img)}} \\
\cmidrule(lr){6-7}
& \smash{\raisebox{0.5ex}{\textbf{(M)}}} & & & & \scriptsize\textbf{\ \ \ \ \ CPU} & \scriptsize\textbf{\ GPU} \\
\midrule
\multicolumn{5}{l}{\textit{(a) Backbone}} & & \\ 
\midrule
ResNet-18 & \textbf{3.6} & 94.4 & 88.7 & 61.5& \textbf{\ \ \ \ 11.8} & \textbf{\ 0.4} \\
Inception-v3 & 4.6 & 95.5& 92.2& 69.4& \ \ \ \ 29.6 & \ 1.0 \\
Inc-Res-v2 \checkmark & 5.5 & \textbf{95.7} & \textbf{93.5} & \textbf{75.6}& \ \ \ \ 41.7 & \ 1.4 \\
\midrule
\multicolumn{5}{l}{\textit{(b) Downsampling Stage}} & & \\ 
\midrule
Second & \textbf{3.4}& 85.9& 88.5& 45.0& \ \ \ \ 42.4& \ 2.0\\
Third \checkmark & 5.5 & \textbf{95.7} & 93.5 & \textbf{75.6}& \ \ \ \ \textbf{41.7} & \ \textbf{1.4} \\
Fourth & 34.3& 95.6& \textbf{95.5}& 70.0& \ \ \ \ 89.6& \ 3.1\\
\midrule
\multicolumn{5}{l}{\textit{(c) Convolutional Blocks}} & & \\ 
\midrule
1 & \textbf{3.2} & 88.5 & 90.5 & 71.5  & \ \ \ \ \textbf{36.4} & \ \textbf{1.1}\\
\textbf{2 \checkmark}  & 5.5 & \textbf{95.7} & \textbf{93.5} & \textbf{75.6}& \ \ \ \ 41.7 & \ 1.4 \\
3 & 7.9& 94.0& 90.5 & 71.5  & \ \ \ \ 46.3& \ 1.6\\
\midrule
\multicolumn{5}{l}{\textit{(d) Balanced Training}} & & \\ 
\midrule
No & 5.5 & 93.5& 90.8& 73.9& \ \ \ \ 41.7 & \ 1.4 \\
Yes \checkmark & \textbf{5.5} & \textbf{95.7} & \textbf{93.5} & \textbf{75.6}& \ \ \ \ \textbf{41.7} & \ \textbf{1.4} \\
\bottomrule
\end{tabular}
\end{table}

\subsubsection{Backbone:} We compare three pre-trained backbones, ResNet-18, Inception-v3, and Inception-ResNet-v2 (Table \ref{tab:ablation_study}(a)). While ResNet-18 offers the fastest inference times (11.8 ms CPU, 0.4 ms GPU), Inception-ResNet-v2 achieves the highest detection and localisation metrics (95.7\% I-AUROC, 93.5\% P-AUROC, 75.6\% AUPRO). Despite its higher inference time (41.7 ms CPU, 1.4 ms GPU), the modest parameter increase from just 3.6 M to 5.5 M justifies the substantial performance gains, making it a justifiable backbone choice.

\subsubsection{Downsampling Stage:} We compare the performance of the Inception-ResNet-v2 backbone built up to the second, third and fourth downsampling stages (Table \ref{tab:ablation_study}(b)). With only two stages, performance drops significantly (85.9\% I-AUROC, 88.5\% P-AUROC, 45.0\% AUPRO) due to the insufficient extraction of high-level features. Extending to the fourth stage produces richer features and improves AUROC (95.6\% I-AUROC, 95.5\% P-AUROC), but causes a substantial increase in parameters (34.3 M) while producing heatmaps that are too small, and resizing such small maps reduces spatial precision and lowers AUPRO to 70.0\%. The third-stage backbone balances rich feature extraction and anomaly heatmap reconstruction (95.7\% I-AUROC, 93.5\% P-AUROC, 75.6\% AUPRO), and therefore is the preferred choice. 

\subsubsection{Convolutional Blocks:} We analyse the effect of varying the number of convolutional blocks appended to the backbone (Table \ref{tab:ablation_study}(c)). The model underperforms while using only a single block (88.5\% I-AUROC, 90.5\% P-AUROC, 71.5\% AUPRO), indicating insufficient refinement of high-level features. Increasing the number of blocks to two improves all metrics (95.7\% I-AUROC, 93.5\% P-AUROC, 75.6\% AUPRO), demonstrating that an additional capacity is beneficial. However, adding a third block reduces performance (94.0\% I-AUROC, 90.5\% P-AUROC, 71.5\% AUPRO), indicating overfitting due to excess model capacity. Therefore, two convolutional blocks provide the best balance between accuracy, model complexity, and efficiency.

\subsubsection{Balanced Training:}

We analyse the effect of balanced sampling during training (Table \ref{tab:ablation_study}(d)). Applying balanced sampling mitigates bias towards more frequent classes such as normal images, stabilises training and improves performance across all metrics (95.7\% I-AUROC, 93.5\% P-AUROC, 75.6\% AUPRO) compared to training without it, thereby justifying its adoption.

\section{Conclusion}

In this work, we proposed MultiTypeFCDD, a lightweight, unified, and explainable convolutional framework for multi-type anomaly detection. Our key contribution is a weakly supervised model that, using only image-level labels, can simultaneously detect anomalies across multiple object categories and, crucially, differentiate their specific types. 
Furthermore, in contrast to massive VLMs that are computationally and memory-intensive, our model is designed as a lightweight and practical alternative, making it suitable for real-world, resource-constrained systems. We demonstrated on the Real-IAD dataset that our framework retains competitive performance against complex state-of-the-art methods while significantly reducing parametric load and inference times. 

\bibliographystyle{splncs04}
\bibliography{bibliography}

@inproceedings{liznerski_explainable_2021,
title={Explainable Deep One-Class Classification},
author={Philipp Liznerski and Lukas Ruff and Robert A. Vandermeulen and Billy Joe Franks and Marius Kloft and Klaus Robert Muller},
booktitle={International Conference on Learning Representations},
year={2021},
}

@article{alzarooni2025anomaly,
  title={Anomaly detection for industrial applications, its challenges, solutions, and future directions: A review},
  author={Alzarooni, Abdelrahman and Iqbal, Ehtesham and Khan, Samee Ullah and Javed, Sajid and Moyo, Brain and Abdulrahman, Yusra},
  journal={arXiv preprint arXiv:2501.11310},
  year={2025}
}

@article{pang_2021_deep_anomaly_detection_survey,
  title={Deep learning for anomaly detection: A review},
  author={Pang, Guansong and Shen, Chunhua and Cao, Longbing and Hengel, Anton Van Den},
  journal={ACM computing surveys (CSUR)},
  volume={54},
  number={2},
  pages={1--38},
  year={2021},
  publisher={ACM New York, NY, USA}
}

@ARTICLE{ruff_2021_anomaly_review,
  author={Ruff, Lukas and Kauffmann, Jacob R. and Vandermeulen, Robert A. and Montavon, Grégoire and Samek, Wojciech and Kloft, Marius and Dietterich, Thomas G. and Müller, Klaus-Robert},
  journal={Proceedings of the IEEE}, 
  title={A Unifying Review of Deep and Shallow Anomaly Detection}, 
  year={2021},
  volume={109},
  number={5},
  pages={756-795}}

@inproceedings{bogdoll_2022_anomaly_driving_survey,
  title={Anomaly detection in autonomous driving: A survey},
  author={Bogdoll, Daniel and Nitsche, Maximilian and Z{\"o}llner, J Marius},
  booktitle={Proceedings of the IEEE/CVF conference on computer vision and pattern recognition},
  pages={4488--4499},
  year={2022}
}

@article{li2023survey,
  title={A survey on explainable anomaly detection},
  author={Li, Zhong and Zhu, Yuxuan and Van Leeuwen, Matthijs},
  journal={ACM Transactions on Knowledge Discovery from Data},
  volume={18},
  number={1},
  pages={1--54},
  year={2023},
  publisher={ACM New York, NY}
}

@inproceedings{defard2021padim,
  title={Padim: a patch distribution modeling framework for anomaly detection and localization},
  author={Defard, Thomas and Setkov, Aleksandr and Loesch, Angelique and Audigier, Romaric},
  booktitle={International conference on pattern recognition},
  pages={475--489},
  year={2021},
  organization={Springer}
}

@inproceedings{liu2023simplenet,
  title={Simplenet: A simple network for image anomaly detection and localization},
  author={Liu, Zhikang and Zhou, Yiming and Xu, Yuansheng and Wang, Zilei},
  booktitle={Proceedings of the IEEE/CVF conference on computer vision and pattern recognition},
  pages={20402--20411},
  year={2023}
}

@inproceedings{roth2022towards,
  title={Towards total recall in industrial anomaly detection},
  author={Roth, Karsten and Pemula, Latha and Zepeda, Joaquin and Sch{\"o}lkopf, Bernhard and Brox, Thomas and Gehler, Peter},
  booktitle={Proceedings of the IEEE/CVF conference on computer vision and pattern recognition},
  pages={14318--14328},
  year={2022}
}

@article{foorthuis2021nature,
  title={On the nature and types of anomalies: a review of deviations in data},
  author={Foorthuis, Ralph},
  journal={International journal of data science and analytics},
  volume={12},
  number={4},
  pages={297--331},
  year={2021},
  publisher={Springer}
}

@inproceedings{mokhtar2025detect,
  title={Detect, Classify, Act: Categorizing Industrial Anomalies with Multi-Modal Large Language Models},
  author={Mokhtar, Sassan and Mousakhan, Arian and Galesso, Silvio and Tayyub, Jawad and Brox, Thomas},
  booktitle={Proceedings of the Computer Vision and Pattern Recognition Conference},
  pages={4058--4067},
  year={2025}
}

@article{sadikaj2025multiads,
  title={MultiADS: Defect-aware Supervision for Multi-type Anomaly Detection and Segmentation in Zero-Shot Learning},
  author={Sadikaj, Ylli and Zhou, Hongkuan and Halilaj, Lavdim and Schmid, Stefan and Staab, Steffen and Plant, Claudia},
  journal={arXiv preprint arXiv:2504.06740},
  year={2025}
}

@inproceedings{gu2024anomalygpt,
  title={Anomalygpt: Detecting industrial anomalies using large vision-language models},
  author={Gu, Zhaopeng and Zhu, Bingke and Zhu, Guibo and Chen, Yingying and Tang, Ming and Wang, Jinqiao},
  booktitle={Proceedings of the AAAI conference on artificial intelligence},
  volume={38},
  number={3},
  pages={1932--1940},
  year={2024}
}

@article{papa2024survey,
  title={A survey on efficient vision transformers: algorithms, techniques, and performance benchmarking},
  author={Papa, Lorenzo and Russo, Paolo and Amerini, Irene and Zhou, Luping},
  journal={IEEE transactions on pattern analysis and machine intelligence},
  volume={46},
  number={12},
  pages={7682--7700},
  year={2024},
  publisher={IEEE}
}

@article{garcia2019estimation,
  title={Estimation of energy consumption in machine learning},
  author={Garc{\'\i}a-Mart{\'\i}n, Eva and Rodrigues, Crefeda Faviola and Riley, Graham and Grahn, H{\aa}kan},
  journal={Journal of Parallel and Distributed Computing},
  volume={134},
  pages={75--88},
  year={2019},
  publisher={Elsevier}
}

@inproceedings{strubell2020energy,
  title={Energy and policy considerations for modern deep learning research},
  author={Strubell, Emma and Ganesh, Ananya and McCallum, Andrew},
  booktitle={Proceedings of the AAAI conference on artificial intelligence},
  volume={34},
  number={09},
  pages={13693--13696},
  year={2020}
}

@article{marshall2025transformers,
  title={Are transformers truly foundational for robotics?},
  author={Marshall, James AR and Barron, Andrew B},
  journal={npj Robotics},
  volume={3},
  number={1},
  pages={9},
  year={2025},
  publisher={Nature Publishing Group UK London}
}

@article{scholkopf2001estimating,
  title={Estimating the support of a high-dimensional distribution},
  author={Sch{\"o}lkopf, Bernhard and Platt, John C and Shawe-Taylor, John and Smola, Alex J and Williamson, Robert C},
  journal={Neural computation},
  volume={13},
  number={7},
  pages={1443--1471},
  year={2001},
  publisher={MIT Press}
}

@inproceedings{liu2008isolation,
  title={Isolation forest},
  author={Liu, Fei Tony and Ting, Kai Ming and Zhou, Zhi-Hua},
  booktitle={2008 eighth ieee international conference on data mining},
  pages={413--422},
  year={2008},
  organization={IEEE}
}

@article{tax2004support,
  title={Support vector data description},
  author={Tax, David MJ and Duin, Robert PW},
  journal={Machine learning},
  volume={54},
  number={1},
  pages={45--66},
  year={2004},
  publisher={Springer}
}

@inproceedings{ruff2018deep,
  title={Deep one-class classification},
  author={Ruff, Lukas and Vandermeulen, Robert and Goernitz, Nico and Deecke, Lucas and Siddiqui, Shoaib Ahmed and Binder, Alexander and M{\"u}ller, Emmanuel and Kloft, Marius},
  booktitle={International conference on machine learning},
  pages={4393--4402},
  year={2018},
  organization={PMLR}
}

@article{arrieta2020explainable,
  title={Explainable Artificial Intelligence (XAI): Concepts, taxonomies, opportunities and challenges toward responsible AI},
  author={Arrieta, Alejandro Barredo and D{\'\i}az-Rodr{\'\i}guez, Natalia and Del Ser, Javier and Bennetot, Adrien and Tabik, Siham and Barbado, Alberto and Garc{\'\i}a, Salvador and Gil-L{\'o}pez, Sergio and Molina, Daniel and Benjamins, Richard and others},
  journal={Information fusion},
  volume={58},
  pages={82--115},
  year={2020},
  publisher={Elsevier}
}

@article{bergmann2021mvtec,
  title={The MVTec anomaly detection dataset: a comprehensive real-world dataset for unsupervised anomaly detection},
  author={Bergmann, Paul and Batzner, Kilian and Fauser, Michael and Sattlegger, David and Steger, Carsten},
  journal={International Journal of Computer Vision},
  volume={129},
  number={4},
  pages={1038--1059},
  year={2021},
  publisher={Springer}
}

@inproceedings{zou2022spot,
  title={Spot-the-difference self-supervised pre-training for anomaly detection and segmentation},
  author={Zou, Yang and Jeong, Jongheon and Pemula, Latha and Zhang, Dongqing and Dabeer, Onkar},
  booktitle={European conference on computer vision},
  pages={392--408},
  year={2022},
  organization={Springer}
}

@article{you2022unified,
  title={A unified model for multi-class anomaly detection},
  author={You, Zhiyuan and Cui, Lei and Shen, Yujun and Yang, Kai and Lu, Xin and Zheng, Yu and Le, Xinyi},
  journal={Advances in Neural Information Processing Systems},
  volume={35},
  pages={4571--4584},
  year={2022}
}

@article{he2024mambaad,
  title={Mambaad: Exploring state space models for multi-class unsupervised anomaly detection},
  author={He, Haoyang and Bai, Yuhu and Zhang, Jiangning and He, Qingdong and Chen, Hongxu and Gan, Zhenye and Wang, Chengjie and Li, Xiangtai and Tian, Guanzhong and Xie, Lei},
  journal={Advances in Neural Information Processing Systems},
  volume={37},
  pages={71162--71187},
  year={2024}
}

@inproceedings{jeong2023winclip,
  title={Winclip: Zero-/few-shot anomaly classification and segmentation},
  author={Jeong, Jongheon and Zou, Yang and Kim, Taewan and Zhang, Dongqing and Dabeer, Onkar},
  booktitle={Proceedings of the IEEE/CVF Conference on Computer Vision and Pattern Recognition},
  pages={19606--19616},
  year={2023}
}

@article{jiang2023weakly,
  title={Weakly supervised anomaly detection: A survey},
  author={Jiang, Minqi and Hou, Chaochuan and Zheng, Ao and Hu, Xiyang and Han, Songqiao and Huang, Hailiang and He, Xiangnan and Yu, Philip S and Zhao, Yue},
  journal={arXiv preprint arXiv:2302.04549},
  year={2023}
}

@article{yang2024anomaly,
  title={Anomaly-guided weakly supervised lesion segmentation on retinal OCT images},
  author={Yang, Jiaqi and Mehta, Nitish and Demirci, Gozde and Hu, Xiaoling and Ramakrishnan, Meera S and Naguib, Mina and Chen, Chao and Tsai, Chia-Ling},
  journal={Medical image analysis},
  volume={94},
  pages={103139},
  year={2024},
  publisher={Elsevier}
}

@inproceedings{wang2024real,
  title={Real-iad: A real-world multi-view dataset for benchmarking versatile industrial anomaly detection},
  author={Wang, Chengjie and Zhu, Wenbing and Gao, Bin-Bin and Gan, Zhenye and Zhang, Jiangning and Gu, Zhihao and Qian, Shuguang and Chen, Mingang and Ma, Lizhuang},
  booktitle={Proceedings of the IEEE/CVF Conference on Computer Vision and Pattern Recognition},
  pages={22883--22892},
  year={2024}
}

@article{chen2025can,
  title={Can multimodal large language models be guided to improve industrial anomaly detection?},
  author={Chen, Zhiling and Chen, Hanning and Imani, Mohsen and Imani, Farhad},
  journal={arXiv preprint arXiv:2501.15795},
  year={2025}
}

@inproceedings{gudovskiy2022cflow,
  title={Cflow-ad: Real-time unsupervised anomaly detection with localization via conditional normalizing flows},
  author={Gudovskiy, Denis and Ishizaka, Shun and Kozuka, Kazuki},
  booktitle={Proceedings of the IEEE/CVF winter conference on applications of computer vision},
  pages={98--107},
  year={2022}
}

@article{jiang2022softpatch,
  title={Softpatch: Unsupervised anomaly detection with noisy data},
  author={Jiang, Xi and Liu, Jianlin and Wang, Jinbao and Nie, Qiang and Wu, Kai and Liu, Yong and Wang, Chengjie and Zheng, Feng},
  journal={Advances in Neural Information Processing Systems},
  volume={35},
  pages={15433--15445},
  year={2022}
}

@inproceedings{zhang2023destseg,
  title={Destseg: Segmentation guided denoising student-teacher for anomaly detection},
  author={Zhang, Xuan and Li, Shiyu and Li, Xi and Huang, Ping and Shan, Jiulong and Chen, Ting},
  booktitle={Proceedings of the IEEE/CVF conference on computer vision and pattern recognition},
  pages={3914--3923},
  year={2023}
}

@inproceedings{deng2022anomaly,
  title={Anomaly detection via reverse distillation from one-class embedding},
  author={Deng, Hanqiu and Li, Xingyu},
  booktitle={Proceedings of the IEEE/CVF conference on computer vision and pattern recognition},
  pages={9737--9746},
  year={2022}
}

@article{fernando2021deep,
  title={Deep learning for medical anomaly detection--a survey},
  author={Fernando, Tharindu and Gammulle, Harshala and Denman, Simon and Sridharan, Sridha and Fookes, Clinton},
  journal={ACM Computing Surveys (CSUR)},
  volume={54},
  number={7},
  pages={1--37},
  year={2021},
  publisher={ACM New York, NY, USA}
}

@article{shank2013coupon,
  title={Coupon collector problem for non-uniform coupons and random quotas},
  author={Shank, Nathan B and Yang, Hannah},
  journal={the electronic journal of combinatorics},
  pages={P33--P33},
  year={2013}
}

@inproceedings{szegedy2017inception,
  title={Inception-v4, inception-resnet and the impact of residual connections on learning},
  author={Szegedy, Christian and Ioffe, Sergey and Vanhoucke, Vincent and Alemi, Alexander},
  booktitle={Proceedings of the AAAI conference on artificial intelligence},
  volume={31},
  number={1},
  year={2017}
}

@inproceedings{deng2009imagenet,
  title={Imagenet: A large-scale hierarchical image database},
  author={Deng, Jia and Dong, Wei and Socher, Richard and Li, Li-Jia and Li, Kai and Fei-Fei, Li},
  booktitle={2009 IEEE conference on computer vision and pattern recognition},
  pages={248--255},
  year={2009},
  organization={Ieee}
}

\end{document}